\documentclass[journal]{IEEEtran}
\usepackage{amsmath,amsfonts}
\usepackage{algorithmic}
\usepackage{array}
\usepackage[caption=false,font=normalsize,labelfont=sf,textfont=sf]{subfig}
\usepackage{textcomp}
\usepackage{stfloats}
\usepackage{url}
\usepackage{verbatim}
\usepackage{graphicx}
\def\BibTeX{{\rm B\kern-.05em{\sc i\kern-.025em b}\kern-.08em
    T\kern-.1667em\lower.7ex\hbox{E}\kern-.125emX}}
\usepackage{balance}

\usepackage[numbers]{natbib}
\usepackage{multicol}
\usepackage{multirow} 
\usepackage{bm}
\usepackage{booktabs}
\usepackage[table]{xcolor}
\usepackage[scr=boondox,cal=esstix]{mathalpha}

\usepackage{color}
\newcommand{\tabincell}[2]{\begin{tabular}{@{}#1@{}}#2\end{tabular}}
\newcommand{\etal}{\textit{et al}.}

\begin{document}

\title{Asymmetrical Siamese Network for Point Clouds Normal Estimation}

\author{
~\IEEEauthorblockN{Wei Jin, Jun Zhou, Nannan Li, Haba Madeline, Xiuping Liu} \\
\thanks{The authors would like to thank the High Performance Computing Center of Dalian Maritime University for providing the computing resources. This research was supported in part by the Natural Science Foundation of China under Grants 62002040, 61976040, and 62201020, in part by China Postdoctoral Science Foundation 2021M690501, and in part by Fundamental Research Funds for the Central Universities (No.~3132024271).(Corresponding author: Jun Zhou.)} 
\thanks{W. Jin is with the Dalian Neusoft University of Information, Dalian 116024, China. (E-mail: jinwei@neusoft.edu.cn). }
\thanks{J. Zhou, N. Li, and H. Madeline are with the School of Information Science and Technology, Dalian Maritime University, Dalian, China (E-mail: zj.9004@gmail.com, nannanli@dlmu.edu.cn, habamadeline@yahoo.com). }
\thanks{X. Liu is with the School of Mathematical Sciences, Dalian University of Technology, Dalian 116024, China. (E-mail: xpLiu@dlut.edu.cn). }
}


\maketitle

\begin{abstract}
In recent years, deep learning-based point cloud normal estimation has made great progress. However, existing methods mainly rely on the PCPNet dataset, leading to overfitting. In addition, the correlation between point clouds with different noise scales remains unexplored, resulting in poor performance in cross-domain scenarios. In this paper, we explore the consistency of intrinsic features learned from clean and noisy point clouds using an Asymmetric Siamese Network architecture. By applying reasonable constraints between features extracted from different branches, we enhance the quality of normal estimation. Moreover, we introduce a novel multi-view normal estimation dataset that includes a larger variety of shapes with different noise levels. Evaluation of existing methods on this new dataset reveals their inability to adapt to different types of shapes, indicating a degree of overfitting. Extensive experiments show that the proposed dataset poses significant challenges for point cloud normal estimation and that our feature constraint mechanism effectively improves upon existing methods and reduces overfitting in current architectures.
\end{abstract}


\section{Introduction}\label{sec:1}
Point normal estimation is a fundamental task in the digital geometry processing community. The estimated normals can be applied to a various downstream fields such as surface reconstruction \cite{berger2014state, kazhdan2006poisson, hashimoto2019normal, kazhdan2013screened, huang2023neural}, point cloud denoising \cite{zhang2020pointfilter}, and semantic segmentation \cite{grilli2017review, che2018multi}. Since traditional methods lack universal parameter settings that can be applied to diverse point clouds, data-driven estimation methods have emerged as a promising research direction to tackle this issue, achieving remarkable results.

Currently, the mainstream normal estimation architectures~\cite{zhu2021adafit, zhou2023improvement, guerrero2018pcpnet, ben2020deepfit, li2022hsurf, li2022graphfit} based on deep learning are mainly inspired by PointNet~\cite{qi2017pointnet} or the subsequent proposed network architectures~\cite{qi2017pointnet++, wang2019dynamic}. These architectures typically consist of three main parts: local patch alignment including PCA preprocessing or quaternion spatial transformer, feature extraction, and normal estimation including direct regression or fitting a local geometric surface. Tab.~\ref{tab:1} provides more details about the comparisons between different architectures. Generally, except for PCA operations, effective feature representation extraction is necessary for each part of the architectures. For instance, in the quaternion spatial transformer operation within the first part, a global feature is extracted to estimate an accurate rigid rotation transformation for input space calibration. Similarly, in the normal estimation part, a global representation is also needed for direct normal estimation or point-wise weight prediction. Therefore, the performance of the networks significantly affects the accuracy and generalization ability of the normal estimation task. In this paper, we propose a novel Asymmetrical Siamese Network for normal estimation, which can be broadly applied on top of existing models such as DeepFit~\cite{lu2020deep} and AdaFit~\cite{zhu2021adafit}. Our Asymmetrical Siamese Network aims to align the multi-stage global features extracted from clean and noisy patches as closely as possible. This way, features from noisy patches can be enhanced by incorporating local geometric priors extracted from clean point clouds. Consequently, the learned features extracted by the networks can be more informative compared to the original models. By using prior information as guidance, we can improve the accuracy of normal estimation.

\begin{table*}[h]
\caption{A comparison of various normal estimation methods in the current mainstream literature.}\label{tab:1}
\footnotesize\
\centering
\begin{tabular}{lcccc}
\toprule
Method & Canonical Pose Estimation & Feature Extraction & Normal Prediction  & Loss\\ 
\midrule
\midrule
PCPNet~\cite{guerrero2018pcpnet} & {\tabincell{c}{PCA\\QSTN}} & PointNet-like & {\tabincell{c}{FC Layers \\(Regression)}} & Euclidean Distance\\
\midrule
Zhou~\etal~\cite{zhou2020normal} & {\tabincell{c}{PCA\\QSTN}} & {\tabincell{c}{PointNet-like\\Multi-scale Expert}} & {\tabincell{c}{FC layers \\(Regression)}}& {\tabincell{c}{Euclidean Distance \\ Cross-entropy Loss}}\\
\midrule
Cao~\etal~\cite{cao2021latent}   & {\tabincell{c}{PCA\\QSTN}} & {\tabincell{c}{PointNet-like\\Multi-scale Expert}} & {\tabincell{c}{Differentiable RANSAC-Like Module \\(Tangent Plane Fitting)}} & Euclidean Distance\\
\midrule 
DeepFit~\cite{ben2020deepfit} &{\tabincell{c}{PCA\\QSTN}} & PointNet-like & {\tabincell{c}{Weight-based Least Squares\\ ($n$-order Surface Fitting)}}& {\tabincell{c}{$sin$ Loss\\Consistency Loss}}\\
\midrule
Zhang~\etal~\cite{zhang2022geometry} &  {\tabincell{c}{PCA\\QSTN}}    &  PointNet-like &  {\tabincell{c}{Weight-based Least Squares \\ Weight-based FC Layers \\ ($n$-order Surface Fitting \& Regression)}}&  {\tabincell{c}{$sin$ Loss\\Euclidean Distance\\Weight-guiding loss}}\\
\midrule
AdaFit~\cite{zhu2021adafit} &{\tabincell{c}{PCA\\QSTN}} &{\tabincell{c}{PointNet-like\\Cascaded Scale Aggregation}} & {\tabincell{c}{Offset Prediction Module \\Weight-based Least Squares\\ ($n$-order Surface Fitting)}} & {\tabincell{c}{$sin$ Loss\\Consistency Loss}}\\
\midrule
GraphFit~\cite{li2022graphfit} &{\tabincell{c}{PCA\\QSTN}} & {\tabincell{c}{DGCNN-like\\Multi-scale Aggregation\\Multi-layer Adaptive Selection}} & {\tabincell{c}{Offset Prediction Module \\Weight-based Least Squares\\($n$-order Surface Fitting)}} & {\tabincell{c}{$sin$ Loss\\Consistency Loss}}\\
\midrule
Hsurf-Net~\cite{li2022hsurf} & PCA &{\tabincell{c}{DGCNN-like\\Relative Position Encoding\\ Cascaded Scale Aggregation}} & {\tabincell{c}{MLP-based Residual Block \\(Regression)}} &  {\tabincell{c}{$sin$ Loss\\Weight-guiding loss}}\\
\bottomrule
\end{tabular}
\end{table*}

In addition, the existing dataset for training and evaluation mainly comes from PCPNet~\cite{guerrero2018pcpnet}, which contains only a small number of samples and exhibits significant differences in noise and sampling patterns compared to real data. Unlike real scanned samples, PCPNet~\cite{guerrero2018pcpnet} dataset do not contain noise patterns caused by changes in the observed view. This makes it challenging to accurately evaluate the effectiveness of existing normal estimation algorithms. In this paper, we introduce a large-scale Multi-View dataset comprising over 5K scanned partial point clouds, each with point-wise ground truth normals provided. Our original models are selected from the ABC dataset~\cite{koch2019abc} and Mesh denoising dataset~\cite{wang2016mesh}. For each selected 3D model, we randomly render 26 partial point clouds from uniformly distributed virtual camera views on a unit sphere. Additionally, we introduce multi-scale noises for each virtual scanned model to further enhance the realism of the simulated environment. Compared with the performance on the PCPNet dataset~\cite{guerrero2018pcpnet}, the accuracy of estimated normals by the most current methods on our proposed dataset exhibits a certain degree of decline. This suggests that different datasets indeed exhibit differences in noisy patterns, and our proposed dataset poses additional challenges in the task of point cloud normal estimation.

In summary, the contributions of this paper are as follows: 
\begin{itemize}
    \item We propose an asymmetrical training paradigm utilizing an Asymmetrical Siamese Network, which effectively extracts valuable information from clean data, ensuring that patterns under multi-scale noise remain consistent with those from noise-free data. This approach significantly enhances the performance of existing mainstream normal estimation methods.
    \item We construct a large-scale multi-view scanned dataset for point normal estimation. Based on this dataset, our study reveals domain gap issues in existing models and demonstrates that training with our extensive dataset can mitigate the impact of these domain gaps.
\end{itemize}

\section{Related Work}\label{sec:2}
With the rapid development of deep learning, data-driven estimation methods have demonstrated advantages in terms of accuracy and generalization compared to conventional methods. In this section, we review both conventional and learning-based methods for normal estimation. Specifically, we categorize learning-based methods into variants of PCPNet~\citep{guerrero2018pcpnet} and others.

\subsection{Conventional Normal Estimation}
The most classic normal estimation algorithms include principal component analysis (PCA) proposed by Hoppe \etal~\cite{hoppe1992surface} and singular value decomposition (SVD) proposed by Klasing \etal~\cite{klasing2009comparison}. These methods sample fixed-scale neighbors for each point on a given model, fitting a local tangent plane to estimate a normal vector at each point. Due to its simplicity and effectiveness, a wide range of variants based on this paradigm have been proposed subsequently, such as moving least squares (MLS)~\cite{levin1998approximation}, truncated Taylor expansion (Jets)~\cite{cazals2005estimating}. These variants aim to fit a higher-order local surface at larger sampling scales to improve the robustness of normal estimation. Despite impressive progress, they are sensitive to sampling scale and struggle to preserve local details. To address the above problems, Mitra~\etal~\cite{mitra2003estimating} combined local information including noise level, curvature, and sampling density to find a more reasonable sampling radius and provide a theoretical guarantee. Then, Pauly~\etal~\cite{pauly2003shape} assigned a Gaussian weight to each neighbor. Additionally, to preserve sharp features and more local details, the analysis of Voronoi cells or Hough transform are applied to the task of normal estimation~\cite{merigot2010voronoi,amenta1998surface,alliez2007voronoi,boulch2012fast}. In fact, as mentioned above, the performance of traditional methods has been greatly improved by using more heuristic designs, but lots of extra nuisance parameters make it difficult to apply these algorithms to more general application scenarios. In recent years, with the rise of deep learning, data-driven methods have achieved noticeable improvements in robustly estimating point cloud normals.

\subsection{PCPNet and its Variants}
Inspired by PointNet~\cite{qi2017pointnet}, Guerrero~\etal~\cite{guerrero2018pcpnet} proposed a point cloud network called PCPNet, which takes a local patch as input and outputs the estimated normal of the patch center. This architecture can be regarded as a milestone in the community of normal estimation and can be roughly divided into three parts: input patch alignment, point feature extraction and the normal estimation. Based on this paradigm, Zhou~\etal~\cite{zhou2020normal} adopted a multi-scale architecture in the feature extraction stage and introduced an additional constrained loss in the normal estimation stage. Cao~\etal~\cite{cao2021latent} designed a differentiable RANSAC-like module to predict latent tangent planes, which can replace the normal estimation module composed of multiple fully connected layers in the PCPNet architecture. Similarly, DeepFit~\cite{ben2020deepfit} utilized the n-order Jet of weighted points to fit local surfaces and predict the center normal. This estimation module can constrain the space of solutions and lead to good normal estimation quality. Subsequently, to improve the reliability of the n-Jet module, Zhang~\etal~\cite{zhang2022geometry} introduced geometric weight guidance, while Zhou~\etal~\cite{zhou2023improvement} and Zhu~\etal~\cite{zhu2021adafit} considered updating the point positions before fitting the local surfaces. Additionally, Zhu~\etal~\cite{zhu2021adafit} also improved the network's ability to extract features by using a novel layer to aggregate features from multiple neighborhood sizes.
Recently, with the help of more powerful feature extraction networks such as point transformer~\cite{Zhao_2021_ICCV} and graph CNN~\cite{wang2019dynamic}, Hsurf~\cite{li2022hsurf} and GraphFit~\cite{li2022graphfit} have further improved the performance of point normal estimation. In this study, we also conduct research based on the PCPNet paradigm. However, the main difference lies in our study focusing on investigating the correlation between feature representation at different noise levels in point cloud normal estimation, with the aim of improving the model performance within this paradigm.

\subsection{Other Kinds of Methods}
In this section, we also review some other impressive learning-based methods. Among them, Boulch and Marlet~\etal~\cite{boulch2016deep} attempted to construct a transformed Hough space accumulator as the input of 2D convolutional neural networks. This marks the first application of deep learning technology to the task of point normal estimation and demonstrates impressive performance. Then, Nesti-Net~\cite{ben2019nesti} used 3D modified Fisher vectors (3DmFV) for local representation within a mixture-of-experts architecture. Zhou~\etal~\cite{zhou2020geometry} proposed a multi-scale descriptor based on local height-map patch (HMP). Lu~\etal~\cite{lu2020deep} also utilized HMP representation and designed a two-step scheme for normal estimation and filtering iteratively. Refine-Net~\cite{zhou2022refine} introduced a normal refinement network based on multiple feature representations extracted from local height-map and local patch points. Compared with PCPNet and its variants, these methods still follow a point-by-point processing pipeline, which can be time-consuming. To enhance the time efficiency, Lenssen~\etal~\cite{lenssen2020deep} proposed a lightweight graph network to realize weights estimation of local neighborhood iteratively, which allows the entire model as input and fast normal estimation. Zhou~\etal~\cite{zhou2022fast} first devised a patch-by-patch processing pipeline combined with transformer technology for rapid and accurate normal estimation. Recently, inspired by neural implicit representations, NeAF~\cite{li2022neaf} utilized a network to learn a neural angle field, aiding in representing normals implicitly. Although improvements have been made by the above methods, the quantitative evaluation is mainly accomplish on PCPNet dataset, which is not suitable for real environment.

\subsection{Benchmarks for Normal Estimation}
The PCV dataset~\cite{zhang2018multi} comprises 152 synthesized point clouds and 288 real scans. Synthetic shapes are sampled from meshes and include multiple Gaussian noise scales, while real scanned shapes only have computed normals as ground truth. The PCPNet Shape Dataset~\cite{guerrero2018pcpnet} contains 30 point clouds sampled uniformly from meshes with 100,000 points (8 for training and 22 for testing). Additionally, each shape contains three levels of Gaussian noise variants and two varying density modes. Although these datasets provide point-wise ground-truth normals, they lack diverse noise patterns found in real scanning environments. The SceneNN dataset~\cite{hua2016scenenn} includes 76 scenes re-annotated with 40 NYU-D v2 classes collected by a depth camera with ground-truth reconstructed meshes. The Semantic3D dataset~\cite{hackel2017semantic3d} consists of dense point clouds acquired with static terrestrial laser scanners covering various urban outdoor scenes. The NYU Depth v2 dataset~\cite{silberman2012indoor} contains aligned and preprocessed RGBD frames, which can be transformed into point clouds for normal estimation. Although these datasets contain rich environmental noise patterns, they lack ground-truth for normal estimation.

\begin{figure*}[htp!]
 \centering
 \includegraphics[width=1\textwidth]{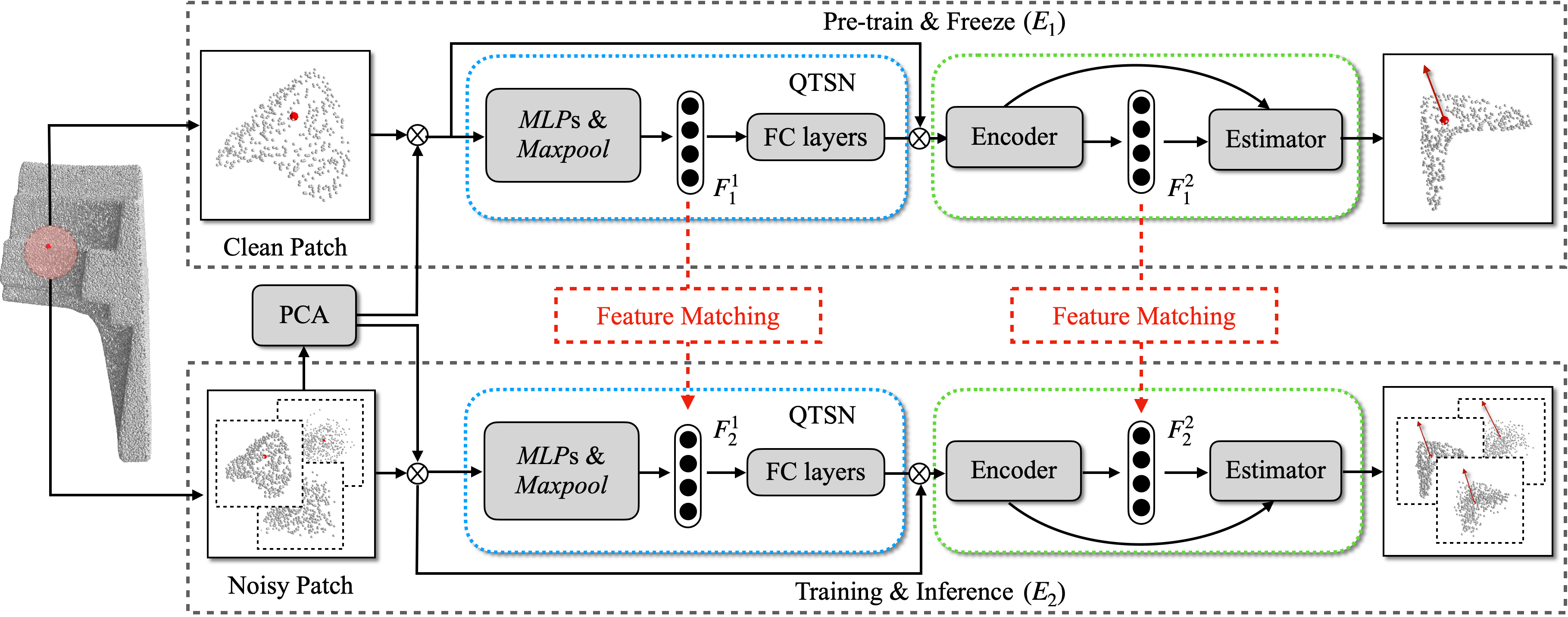}
 \caption{ The architecture of our Asymmetrical Siamese Normal Estimation involves two stages of training. Initially, the upper part of the network is trained using clean, noise-free data. Next, we freeze the weights of the upper part of the network, which now only provides features from the clean data, and train the lower part of the network using complete multi-scale noisy data. The features produced by both branches are constrained to be consistent.In the inference stage, we only use the lower part of the network for normal estimation. }
 \label{fig:2}
\end{figure*}

\section{Network Architecture}\label{sec:3}

\subsection{Overview}
The overall architecture for point normal estimation is illustrated in Fig.\ref{fig:2}. Given a sampled patch as input, we introduce an Asymmetrical Siamese Network to estimate the point normal. Various estimators such as DeepFit\cite{ben2020deepfit} and AdaFit~\cite{zhu2021adafit} can be utilized in this architecture. It maps the noisy patch and clean patch into a shared latent space at different stages of the estimators.

\subsection{Asymmetrical Siamese Network}
To train a normal estimation model less affected by noise, we propose an Asymmetric Siamese Network using a two-stage training strategy, as illustrated in Fig.~\ref{fig:2}. The architecture comprises two normal estimators and a feature metric learning mechanism, with both sub-networks sharing identical architecture. Specifically, the adopted estimators can be built on the paradigm of PCPNet~\cite{guerrero2018pcpnet} or its variants, such as DeepFit\cite{ben2020deepfit} and AdaFit~\cite{zhu2021adafit}. The estimator can generally be divided into three parts: the spatial transform module, the feature encoder module, and the normal prediction module (decoder), as shown in the blue and green dotted boxes, respectively. The spatial transform module, which consists of Multi-layer Perceptrons (MLPs) and fully connected layers, is employed to estimate a quaternion for each input patch and output a predicted canonical patch. Then, for a canonical patch, the encoder and decoder modules perform feature extraction and normal estimation, respectively. Various feature encoder modules can be used, such as MLPs or dynamic graph CNN blocks, while the normal prediction module can be implemented using simple multi-layer fully connected layers or weight-based least squares solvers. Therefore, the used estimator can extract multiple global features associated with the given patch at different stages. By introducing the symmetrical Siamese training mechanism, noise-free information can effectively guide the model training process and improve the model's representation ability for noisy patches.

In addition, for a given patch, to ensure that the two asymmetric branches focus more on intrinsic noise information rather than differences in initial orientation, we apply a shared PCA operation in the preprocessing stage for the two sub-networks. Specifically, we first train the normal estimator $E_{1}$ (shown in the upper part of Fig.~\ref{fig:2}) for noise-free point cloud normal estimation. In this stage, for each noise-free patch, a corresponding noisy patch is required. The PCA operation is performed on the noisy patch to obtain an initial rotation transformation, which is then used for the initial canonical pose estimation of the noise-free patch in branch $E_{1}$. Once the first training process is completed, all weights of the normal estimator $E_{1}$ will be frozen. Subsequently, the second branch $E_{2}$ (the lower part in Fig.~\ref{fig:2}) is trained on noisy patches. The goal is to align the distribution of feature representations  ($F_{1}^{1}$, $F_{2}^{1}$, $F_{1}^{2}$ and $F_{2}^{2}$ denoted in Fig.~\ref{fig:2}) extracted from both branches by optimizing the feature matching distance. Finally, in the inference stage, the second branch $E_{2}$ can perform the normal estimation task independently. Experiments in Sec.~\ref{sec:5.1} demonstrate that the feature representations obtained using the proposed feature matching strategy are more effective than the global features directly extracted from a single branch.

\begin{figure*}[htp]
 \centering 
 \includegraphics[width=0.9\textwidth]{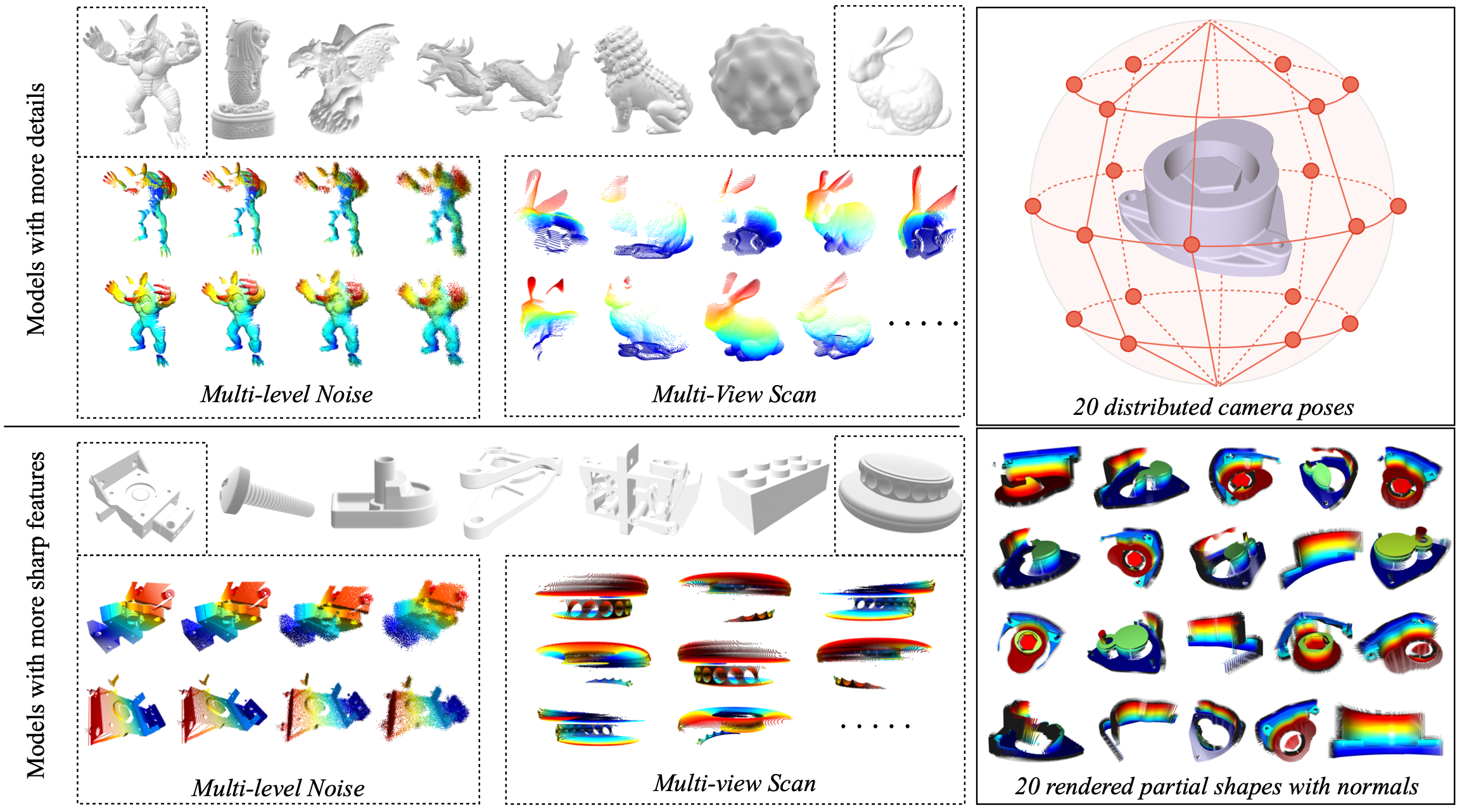}
 \caption{Multi-View Normal Estimation Dataset Construction. This dataset is created using multi-view simulated scanning and introduces noise at different scales. The left side shows examples of multi-view, multi-noise data from CAD models and detailed models, while the right side illustrates the scanning method. }
 \label{fig:1}
\end{figure*}

\subsection{Loss Function}
Our training loss comprises two primary components, which are multi-stage feature matching constraints and normal estimation constraints. The normal estimation constraints typically consist of angle loss and consistency loss, although the specific loss functions may vary slightly across different algorithms.

\noindent
\textbf{Feature matching loss.} To achieve consistency between the multi-stage features of the noisy patch and the noise-free patch, we use the Euclidean distance for feature matching. Thus, the similarity between any two high-dimensional feature vectors can be calculated using the following equation:
\begin{equation}
    L_{diff}^{l} (F_{1}^{l},F_{2}^{l}) = \sum_{j=1}^{C}\|f_{1,j}^{l}-f_{2,j}^{l}\|^{2},
\end{equation}
where $l$ denotes the $l$-th stage of the estimator, $C$ is the dimension of the extracted features. $F_{1}^{l} = (f_{1,1}^{l},f_{1,2}^{l},\cdots,f_{1,C}^{l})^{T}$ and $F_{2}^{l} = (f_{2,1}^{l},f_{2,2}^{l},\cdots,f_{2,C}^{l})^{T}$ represent the extracted features from the noise-free and noisy patches sampled from point clouds.

~

\noindent
\textbf{Normal estimation loss.} Firstly, to predict the point normal that can match the ground-truth as closely as possible, a $sin$ center loss is employed to minimize the distance between the output normal $n_{p}$ and the ground-truth normal $\hat{n}_{p}$ for a query point $p$. This loss has been widely used in mainstream point normal estimation methods~\cite{guerrero2018pcpnet,ben2020deepfit,zhu2021adafit,li2022graphfit,li2022hsurf}, and is defined as: 
\begin{equation}
    L_{center} = \|\hat{n}_{p} \times n_{p} \|.
\end{equation}
It is worth noting that the difference between the estimated normal and the ground-truth can also be quantified using the Euclidean distance as the loss function:
\begin{equation}
    L_{center} = \|\hat{n}_{p} - n_{p} \|^{2}.
\end{equation}
It should be noted that for several architectures based on a weighted least squares solver, a consistency loss is also used to ensure that neighbor points ${p_{i}}\in kNN (q_{i})$ lie on the fitting surface for a query point $q_{i}$, which is defined as:
\begin{equation}
    L_{con} = \frac{1}{N_{q_{i}}}\left [-\lambda_{1}\sum_{j=1}^{N_{p_{i}}}\log(w_{j})+\lambda_{2}\sum_{j=1}^{N_{p_{i}}}w_{j}\|n_{gt,j}\times \hat{n}_{j}\|\right].
\end{equation}
Lastly, for the operation of the QSTN module, a regularization loss $L_{reg} = |I-AA^{T}|$ facilitates much easier optimization, where $A$ is a feature alignment matrix predicted by PointNet. More details can be found in the literature~\cite{guerrero2018pcpnet}. Therefore, the final loss can be written as: 
\begin{equation}
    L_{total} = L_{center}+ \alpha_{1}L_{con}+\alpha_{2}L_{reg}
\end{equation}
where $\alpha_1$ and $\alpha_2$ are hyper-parameters for balancing the effects of all sub-terms on the optimization behavior.

\begin{table*}[h]
\caption{Comparison of the RMSE angle error for unoriented normal estimation of our Asymmetrical Siamese Architecture to the classical deep learning methods on datasets PCPNet Dataset. Here, ``w'' indicates using our method, while ``w/o'' refers to the original methods.}
\normalsize
\centering
\begin{tabular}{l ccc ccc ccc}
\toprule
\multirow{2}{*}{} & \multicolumn{3}{c}{\textbf{DeepFit}} &\multicolumn{3}{c}{\textbf{AdaFit}}& \multicolumn{3}{c}{\textbf{GraphFit}}\\
\cmidrule(r){2-4} \cmidrule(r){5-7}\cmidrule(r){8-10}
&w/o    & w & improv.(\%)   &w/o    & w & improv.(\%)   &w/o    & w & improv.(\%) \\
\midrule
None  &6.51  & 6.04 & 7.24$\uparrow$ &5.19  &5.18  & 0.24$\uparrow$  & 4.45 & 4.56 &-2.53$\downarrow$ \\
\hline
\textbf{Noise} $\sigma$       \\
0.00125 & 9.21  &  9.01 & 2.20$\uparrow$   & 9.05  &8.83  &  2.47$\uparrow$ & 8.74 & 8.52 &2.52$\uparrow$  \\
0.006   & 16.72  & 16.71  & 0.08$\uparrow$ & 16.44 &16.51  &  -0.42$\downarrow$  & 16.05 &16.25  & -1.24$\downarrow$\\
0.012   & 23.12  &  22.59 & 2.28$\uparrow$ & 21.94 & 21.55 & 1.76$\uparrow$  &  21.64 & 21.56 & 0.39$\uparrow$\\
\hline
\textbf{Density}   \\
Gradient & 7.31 & 6.70  &   8.32$\uparrow$ & 5.90 &5.73  & 2.92$\uparrow$& 5.22 & 5.25 & -0.53$\downarrow$\\
Stripes  & 7.92  & 7.31 &   7.69$\uparrow$ & 6.01 & 5.95 & 0.99$\uparrow$ & 5.48 &5.56  & -1.43$\downarrow$\\
\hline
Average  & 11.80 & 11.39 &  3.44$\uparrow$ & 10.76 & 10.62 & 1.22$\uparrow$ & 10.26  & 10.28 & -0.19$\downarrow$\\
\bottomrule
\end{tabular}
\label{tab:3}
\end{table*}

\begin{table*}[h]
\caption{Comparison of the RMSE angle error for unoriented normal estimation of our Asymmetrical Siamese Architecture to the classical deep learning methods on datasets FamousShape Dataset. Here, ``w'' indicates using our method, while ``w/o'' refers to the original methods.}
\normalsize
\centering
\begin{tabular}{l ccc ccc ccc}
\toprule
\multirow{2}{*}{} & \multicolumn{3}{c}{\textbf{DeepFit}} &\multicolumn{3}{c}{\textbf{AdaFit}}& \multicolumn{3}{c}{\textbf{GraphFit}}\\
\cmidrule(r){2-4} \cmidrule(r){5-7}\cmidrule(r){8-10}
&w/o    & w & improv.(\%)   &w/o    & w & improv.(\%)   &w/o    & w & improv.(\%) \\
\midrule
None  &11.21  & 10.56 & 5.80$\uparrow$ &9.09  &9.10  & -0.11$\downarrow$  & 7.88 & 7.82 & 0.76$\uparrow$ \\
\hline
\textbf{Noise} $\sigma$       \\
0.00125 & 16.39  &  15.69 & 4.27$\uparrow$   & 15.78& 15.30 &  3.04$\uparrow$ & 15.78 & 14.82 & 6.08$\uparrow$  \\
0.006   & 29.84  & 29.45 & 1.31$\uparrow$ & 29.78 & 29.40 &  1.28$\uparrow$  & 29.37 &28.90  & 1.60$\uparrow$\\
0.012   & 39.95  &  39.84 & 0.28$\uparrow$ & 38.74 & 38.42& 0.83$\uparrow$  &  38.67& 38.54 &0.34$\uparrow$\\
\hline
\textbf{Density}   \\
Gradient & 10.54 & 9.91  &5.78 $\uparrow$ &8.58 & 8.61& -0.35$\uparrow$& 7.62 & 7.62 & -\\
Stripes  & 11.84  & 10.98 &7.26$\uparrow$ & 8.01 &7.99 & $0.25\uparrow$ & 8.53 &8.59 & -0.70$\downarrow$\\
\hline
Average  & 19.96 & 19.41 &2.76$\uparrow$ & 18.33 & 18.14 & 1.04$\uparrow$ & 17.98 & 17.72 & 1.45 $\uparrow$\\
\bottomrule
\end{tabular}
\label{tab:6}
\end{table*}

\begin{figure*}[htp!]
    \centering
    \includegraphics[width=0.93\textwidth]{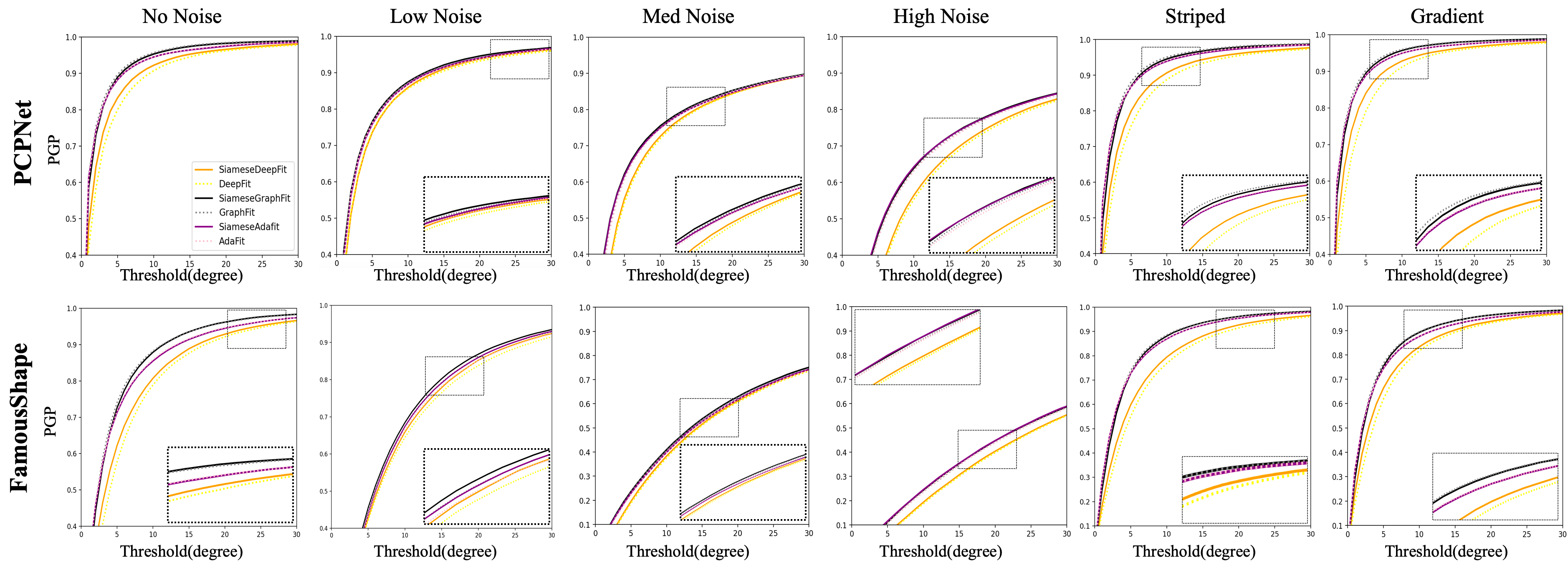}
    \caption{AUC on the PCPNet and FamousShape dataset. X and Y axes are the angle threshold and the percentage of good point (PGP) normals.}
    \label{fig:5}
\end{figure*}

\begin{figure*}[t]
 \centering 
 \includegraphics[width=0.85\textwidth]{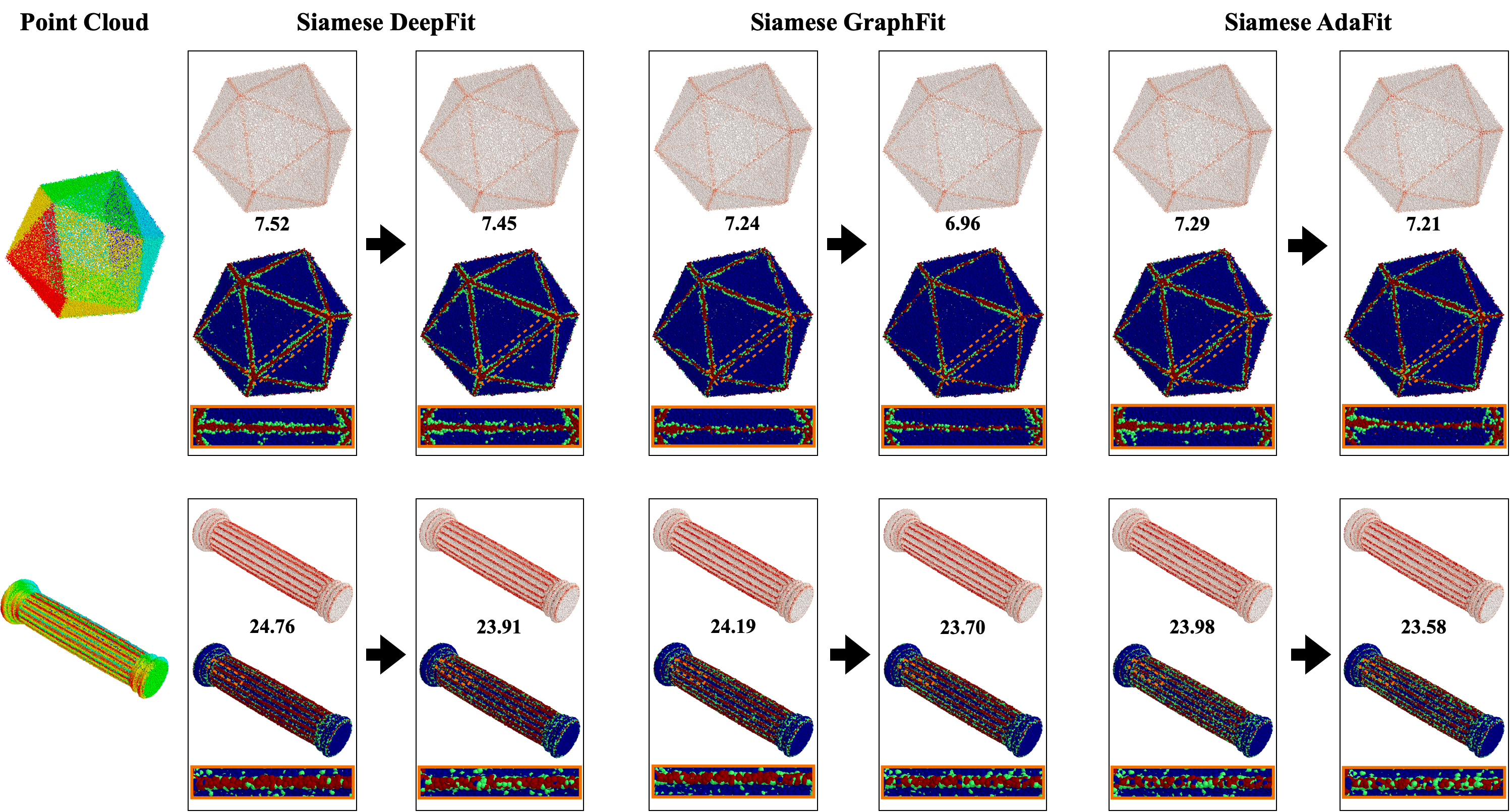}
 \caption{Quantitative analysis of the normal estimation on PCPNet dataset. Upper of each example: the angular error visualization results of our method compared to others. The colors of the points correspond to angular difference, mapped to a heatmap ranging from 0–90 degrees lighter colors indicate better results; Bottom of each example: blue denotes points with an estimated normal that deviates by less than $5^\circ$ from the ground-truth, green encodes angular deviations between $5^\circ$ and $10^\circ$, and red marks errors $>10^\circ$.}
 \label{fig:6}
\end{figure*}

\begin{figure*}[t]
 \centering 
 \includegraphics[width=0.85\textwidth]{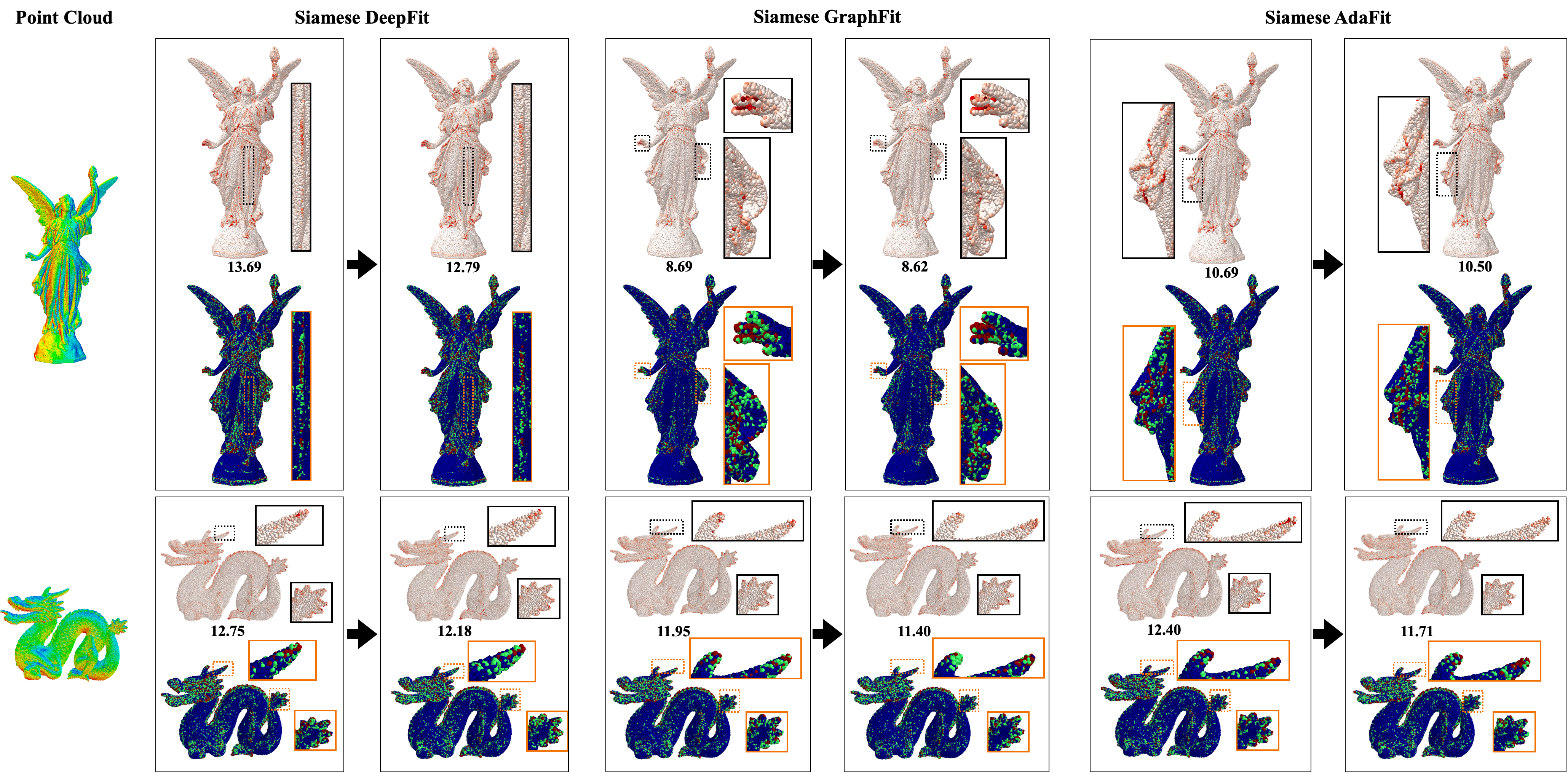}
 \caption{Quantitative analysis of the normal estimation on FamousShape dataset. Upper of each example: the angular error visualization results of our method compared to others. The colors of the points correspond to angular difference, mapped to a heatmap ranging from 0–90 degrees lighter colors indicate better results; Bottom of each example: blue denotes points with an estimated normal that deviates by less than $10^\circ$ from the ground-truth, green encodes angular deviations between $10^\circ$ and $20^\circ$, and red marks errors $>20^\circ$. }
 \label{fig:7}
\end{figure*}

\section{Multi-View Normal Estimation Dataset}\label{sec:4}
Towards an effort to construct a more comprehensive dataset for training and evaluating various methods of point cloud normal estimation, we collected a total of 11 models with rich features from the denoising synthetic dataset~\cite{wang2016mesh} and 56 CAD models from the ABC Dataset~\cite{Koch_2019_CVPR}. For each mesh from the collected dataset, we selected 20 camera poses uniformly distributed on a unit sphere and sampled the rendered point clouds based on these camera poses, as illustrated in the right part of Fig.~\ref{fig:1}. This rendering process was implemented using Blender, with the virtual camera resolution set to $640 \times 480$. Additionally, the width of the sensor and the length of the focal were set to 36 and 40, respectively. Each point of the virtual scanned shape was labeled with the ground-truth normal of the face it was sampled from. Rendering point clouds from various views enhances the diversity of sampling density. 

To simulate the noise that may be introduced during virtual scanning, we added multiple levels of Gaussian noise to the multi-view scanned shapes with standard deviations of $0.0025$, $0.012$, and $0.024$ of the length of the bounding box diagonal of the shape, similar to the setting of PCPNet~\cite{guerrero2018pcpnet}. As illustrated in Fig.~\ref{fig:1}, several examples with different noise levels are displayed. It is important to note that the noise is added to the point positions but no change is made to the ground truth normals, and the points sampled from different noise point clouds have a corresponding relationship.

The proposed multi-view virtual scanned dataset offers several significant advantages: \textbf{1)} It includes both CAD models and synthetic models with more details, enhancing the richness of the normal estimation dataset compared to the PCPNet dataset~\cite{guerrero2018pcpnet}. \textbf{2)} The scanned shapes cover a wide range of sampling densities, and a large variety of shapes are obtained from a relatively small number of 3D mesh models. \textbf{3)} Our dataset comprises a large number of high-quality synthetic partial scans, which accurately simulate real-scanned point clouds affected by self-occlusion.

\section{Experiments}
\subsection{Datasets and Settings}
The datasets used in this study include PCPNet~\cite{guerrero2018pcpnet} and FamousShape~\cite{erler2020points2surf}, both of which provide ground-truth normals for point clouds. The PCPNet dataset consists of two parts: the training set and the test set. We adhere to the same experimental setups, including the train-test split, adding noise, and changing distribution density on the test data. The FamousShape dataset is primarily used to evaluate the performance of the methods. Additionally, this work utilizes the multi-view scanning data proposed in this study. This dataset includes a total of 5360 point clouds with four noise scales, divided into the training set and the test set, with $40\%$ designated as the test set.

\begin{table*}[htp!]
\caption{Comparison of RMSE angle error for normal estimation training, with and without the incorporation of our multi-view dataset, on our detailed models test set from the Multi-View Normal Estimation Dataset. Here, ``w'' indicates the utilization of both our multi-view and PCPNet training datasets, whereas ``w/o'' denotes training solely with the PCPNet training dataset.}
\centering
\resizebox{\textwidth}{!}{\begin{tabular}{l ccc ccc ccc ccc ccc}
\toprule
\multirow{2}{*}{} & \multicolumn{3}{c}{\textbf{PCPNet (single)}}&\multicolumn{3}{c}{\textbf{DeepFit}} &\multicolumn{3}{c}{\textbf{AdaFit}}& \multicolumn{3}{c}{\textbf{GraphFit}}&\multicolumn{3}{c}{\textbf{Hsurf}}\\
\cmidrule(r){2-4} \cmidrule(r){5-7}\cmidrule(r){8-10}\cmidrule(r){11-13}\cmidrule(r){14-16}
&w/o    & w & improv.(\%)  &w/o    & w & improv.(\%)   &w/o    & w & improv.(\%)   &w/o    & w & improv.(\%) &w/o    & w & improv.(\%) \\
\midrule
None    & 11.24  & 8.24   & 26.69$\uparrow$ & 9.01  &  8.22  & 8.77$\uparrow$ & 6.04   &5.61 & 7.12$\uparrow$  & 5.93 & 4.92  & 17.03$\uparrow$  & 5.50 & 5.32 & 3.27$\uparrow$ \\
\hline
\textbf{Noise} $\sigma$       \\
0.00125 &  14.05 & 10.97  & 21.92$\uparrow$ & 11.39  &  10.40 & 7.9$\uparrow$   & 11.79  &9.66  & 18.07$\uparrow$  & 11.25 & 8.96 & 20.36$\uparrow$ & 11.21 & 9.53 & 14.99$\uparrow$  \\
0.006   & 24.40 & 19.11  & 21.68$\uparrow$ & 19.82  & 18.43  & 7.01$\uparrow$ & 21.39  & 18.38  & 14.07$\uparrow$ & 20.11 & 17.36 & 13.67$\uparrow$ & 20.44 & 18.43 & 9.83$\uparrow$ \\
0.012   & 28.28 & 23.77   & 15.95$\uparrow$ & 25.28  &  23.67 & 6.37$\uparrow$  & 25.38 & 23.18 & 8.67$\uparrow$  &  26.04  & 23.51 & 9.72$\uparrow$ & 25.33 & 23.33 & 7.90$\uparrow$ \\
\hline
Average  &  19.49 & 15.52  & 20.37$\uparrow$ & 16.37 & 15.18 & 7.27$\uparrow$  & 16.13 & 14.21 & 11.90$\uparrow$  & 15.83  & 13.69 & 13.52$\uparrow$ & 15.62 & 14.15 & 9.41$\uparrow$ \\
\bottomrule
\end{tabular}
}
\label{tab:4}
\end{table*}

\begin{figure*}[htp]
 \centering 
 \includegraphics[width=0.75\textwidth]{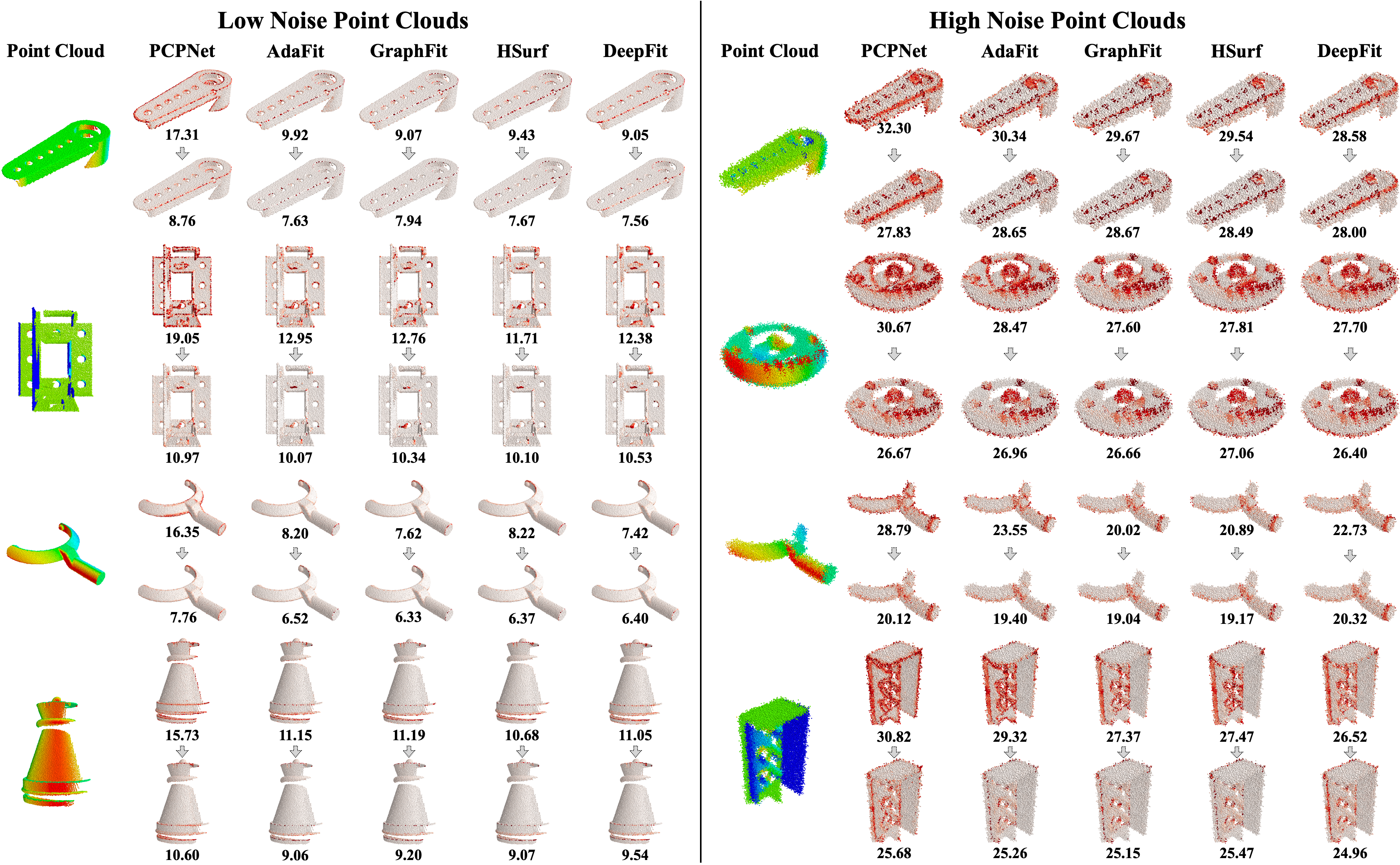}
 \caption{Visual comparison of the improvements brought by training with our multi-view dataset on existing methods, demonstrated on the CAD models from our test set. For each instance, top: without our dataset, bottom: with our dataset.}
 \label{fig:3}
\end{figure*}

\subsection{Enhancements in Asymmetrical Siamese Architecture}\label{sec:5.1}
To validate the effectiveness of our Asymmetrical Siamese framework, we compared it with several mainstream normal estimation algorithms, including DeepFit~\cite{ben2020deepfit}, AdaFit~\cite{zhu2021adafit}, and GraphFit~\cite{li2022graphfit}. To ensure the fairness of the comparison experiments, we used the same parameter settings as in the original works, with the only difference being that we trained two branches using these methods as the backbone. Specifically, we used the noise-free data from the PCPNet training set for the first training stage and the complete PCPNet training set for the second stage of the Asymmetrical Siamese Network. We assessed the accuracy improvements on the PCPNet and FamousShape datasets with and without the Asymmetrical Siamese framework through both quantitative and qualitative comparisons. As shown in Tab.~\ref{tab:3}, our method brings improvements to the DeepFit and AdaFit models on the PCPNet dataset. Although there is a slight decrease in performance for the GraphFit method, our approach consistently enhances performance for data with high noise level across all methods. Then, Tab.~\ref{tab:6} demonstrates that, after training with the Asymmetrical Siamese framework, our method also improves performance on the FamousShape dataset, which is different from the PCPNet dataset.  This indicates that our training approach with two stages can indeed enhance the generalization ability of the models by conveying pattern information from noise-free data. Additionally, the AUC results are illustrated in Fig.~\ref{fig:5}, where our method consistently demonstrates superior performance, indicating its remarkable stability across different angular thresholds. The enlarged part of Fig.~\ref{fig:5} shows that training with our Asymmetric Siamese Network brings noticeable improvements, particularly as the error increases.

Qualitative comparison results on the PCPNet and FamousShape datasets are presented in Fig.~\ref{fig:6} and Fig.~\ref{fig:7}, respectively. To more clearly demonstrate the improvements brought by the Siamese framework, we show the angular error at each point of the point cloud using a heatmap. Additionally, we present the distribution of points with different error ranges. For the PCPNet dataset, we use error ranges of 5 and 10, while for the FamousShape dataset, we use error ranges of 10 and 20. The experiments show that the training approach based on the Siamese structure reduces the occurrence of large errors. This indicates that the feature constraint successfully conveys the characteristics of noise-free to the high-error samples.

\begin{figure*}[htp!]
 \centering 
 \includegraphics[width=0.75\textwidth]{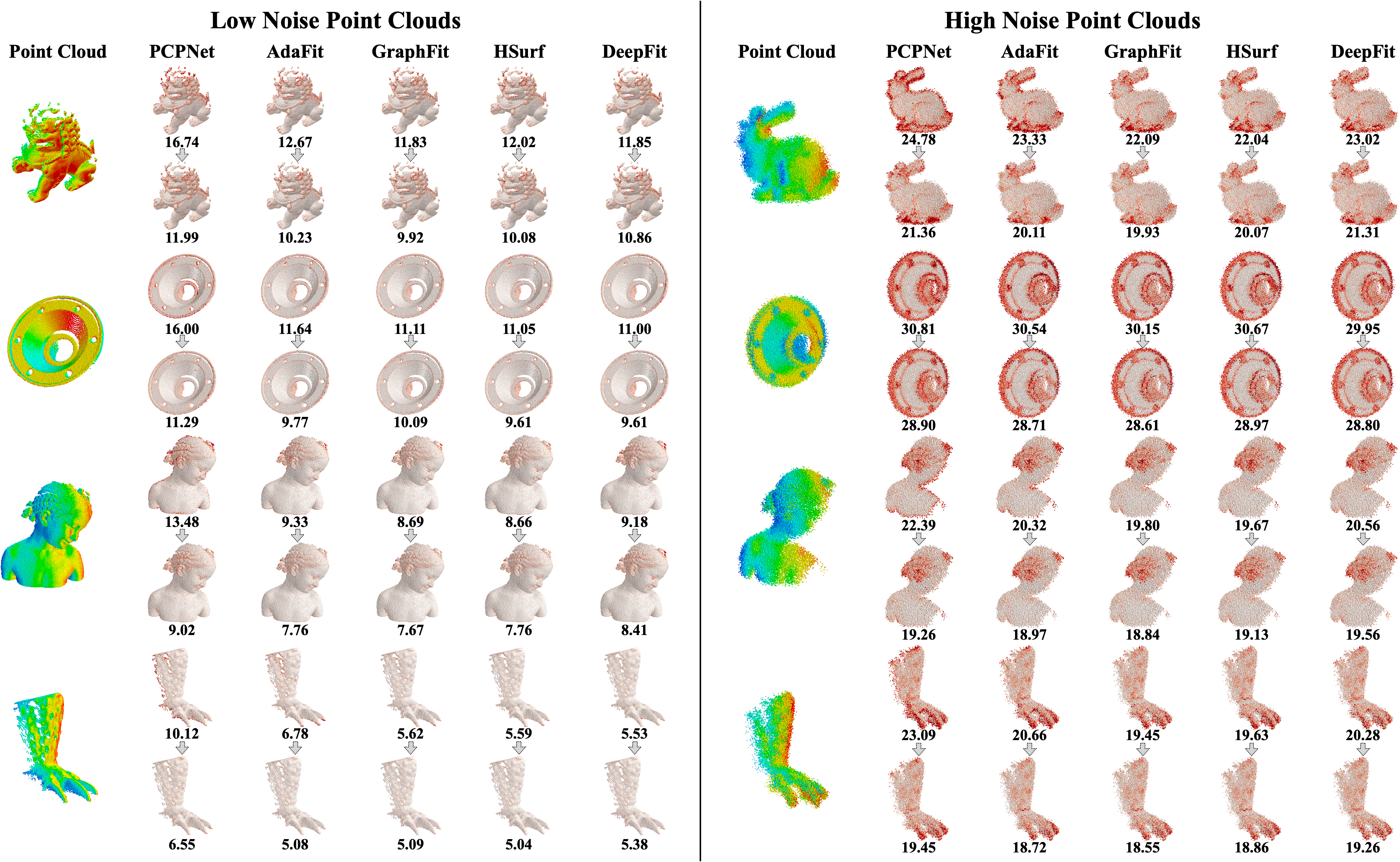}
 \caption{Visual comparison of the improvements brought by training with our multi-view dataset on existing methods, demonstrated on the detailed models from our test set. For each instance, top: without our dataset, bottom: with our dataset.}
 \label{fig:4}
\end{figure*}

\begin{table*}[htp!]
\caption{Comparison of RMSE angle error for normal estimation training on our CAD test set from the Multi-View Normal Estimation Dataset, with and without the incorporation of our multi-view dataset. ``w'' indicates the utilization of both our multi-view and PCPNet training datasets, whereas ``w/o'' denotes training solely with the PCPNet training dataset.}
\centering
\resizebox{\textwidth}{!}{\begin{tabular}{l ccc ccc ccc ccc ccc}
\toprule
\multirow{2}{*}{} & \multicolumn{3}{c}{\textbf{PCPNet (single)}}&\multicolumn{3}{c}{\textbf{DeepFit}} &\multicolumn{3}{c}{\textbf{AdaFit}}& \multicolumn{3}{c}{\textbf{GraphFit}}&\multicolumn{3}{c}{\textbf{Hsurf}}\\
\cmidrule(r){2-4} \cmidrule(r){5-7}\cmidrule(r){8-10}\cmidrule(r){11-13}\cmidrule(r){14-16}
&w/o    & w & improv.(\%)  &w/o    & w & improv.(\%)   &w/o    & w & improv.(\%)   &w/o    & w & improv.(\%) &w/o    & w & improv.(\%)\\
\midrule
None    & 11.72 & 7.30  & 37.71$\uparrow$  & 8.87  & 6.96  & 21.53$\uparrow$  &6.50  & 4.56  & 29.85$\uparrow$  & 5.67  &  5.50 & 3.00$\uparrow$ & 5.77 & 4.39 & 23.92$\uparrow$ \\
\hline
\textbf{Noise} $\sigma$       \\
0.00125 & 11.93 & 10.31   & 13.58$\uparrow$  & 11.42   &  9.56   & 16.29$\uparrow$   & 11.79  & 8.93   &  25.40$\uparrow$  &  11.24 &  9.54 & 15.12$\uparrow$ & 11.21 & 9.53  & 14.99$\uparrow$ \\
0.006   & 25.54 & 18.17   & 28.86$\uparrow$  & 19.84   &  17.63  & 11.14$\uparrow$   & 21.27  & 17.51  &  17.68$\uparrow$  & 20.25  & 18.15  & 10.37$\uparrow$ & 20.04 & 17.61 & 12.13$\uparrow$ \\
0.012   & 30.49 & 24.40   &  19.97$\uparrow$ & 26.15   &  24.12  & 7.76$\uparrow$   & 27.54  & 23.69  &  13.98$\uparrow$  &  25.01 & 23.04  & 7.88$\uparrow$ & 26.38 & 23.84 & 9.63$\uparrow$ \\
\hline
Average  & 20.67 &15.04   & 27.24$\uparrow$  & 16.57   & 14.57   & 12.07$\uparrow$   & 16.77  & 13.67  & 18.49$\uparrow$   & 15.54  &  14.06 & 9.52$\uparrow$ & 15.62 & 14.15 & 9.41$\uparrow$ \\
\bottomrule
\end{tabular}
}
\label{tab:5}
\end{table*}

\begin{table*}[htp!]
\caption{Comparison of RMSE angle error for normal estimation training, with and without the incorporation of our multi-view dataset, on the PCPNet test dataset. Here, ``w'' indicates the utilization of both our multi-view and PCPNet training datasets, whereas ``w/o'' denotes training solely with the PCPNet training dataset.}
\centering
\resizebox{\textwidth}{!}{
\begin{tabular}{l ccc ccc ccc ccc ccc}
\toprule
\multirow{2}{*}{} & \multicolumn{3}{c}{\textbf{PCPNet (single)}}&\multicolumn{3}{c}{\textbf{DeepFit (256)}} &\multicolumn{3}{c}{\textbf{AdaFit}}& \multicolumn{3}{c}{\textbf{GraphFit}}& \multicolumn{3}{c}{\textbf{Hsurf}}\\
\cmidrule(r){2-4} \cmidrule(r){5-7}\cmidrule(r){8-10}\cmidrule(r){11-13}\cmidrule(r){14-16}
&w/o    & w & improv.(\%)  &w/o    & w & improv.(\%)   &w/o    & w & improv.(\%)   &w/o    & w & improv.(\%) &w/o    & w & improv.(\%)\\
\midrule
None    & 9.66 & 8.51  &11.95$\uparrow$  &6.51 &6.67 & -2.43$\downarrow$ &5.19 & 5.23 & -0.77$\downarrow$& 4.61 &5.21&13.02$\downarrow$ & 4.32 & 4.74 &-9.72$\downarrow$ \\
\hline
\textbf{Noise} $\sigma$  \\
0.00125 & 11.46  &10.61  &7.45$\uparrow$   &9.21 &9.23 & -0.26$\downarrow$ &9.05 &9.24 & -2.10$\downarrow$ &8.82&8.72& 1.13$\uparrow$ & 8.95 & 8.96 & -0.11$\downarrow$\\
0.006   &18.26   &17.64  &3.38$\uparrow$   & 16.72& 16.75& -0.19$\downarrow$ &16.44 & 16.55 & -0.67$\downarrow$ & 16.10 & 16.56 & -2.86$\downarrow$ & 16.27 & 16.48 & -1.29$\downarrow$\\
0.012   & 22.80  &22.27  &2.32$\uparrow$  &23.12 & 22.84& 1.23$\uparrow$ &21.94 & 21.93 & 0.05$\uparrow$ &21.69&22.07& -1.75$\downarrow$ & 21.62 & 21.98 & -1.67$\downarrow$ \\
\hline
\textbf{Density}   \\
Gradient & 11.74  &10.11 &24.67$\uparrow$&7.31 & 7.27& 0.58$\uparrow$ &5.90& 5.77 & 2.20$\uparrow$ &5.16&5.82& -12.79$\downarrow$ &5.13&5.46& -6.43$\downarrow$ \\
Stripes  &13.42   &9.71  &17.27$\uparrow$ &7.92 &7.86 &  0.79$\uparrow$  &6.01 &6.07& -1.00$\downarrow$ &5.38&6.00& -11.52$\downarrow$ &4.90&5.24& -6.94$\downarrow$ \\
\hline
Average   &14.56  &13.14  &9.72$\uparrow$&11.80 & 11.77&  0.25$\uparrow$ &10.76 &10.80& -0.37$\downarrow$ &10.29&10.73& -4.28$\downarrow$ &10.20&10.48& -2.75$\downarrow$ \\
\bottomrule 
\end{tabular}
}
\label{tab:2}
\end{table*}

\subsection{Domain Gap Between Datasets}
In this section of the study, we explored the impact of differences between datasets on model performance. Firstly, we evaluated the multi-view test dataset using existing mainstream methods. As shown in Tab.~\ref{tab:5} and Tab.~\ref{tab:4} , results from testing solely with the PCPNet dataset (``w/o'') indicate poorer performance on our proposed dataset, including CAD models and detailed models. This highlights significant differences between the two datasets. However, after retraining existing models with a portion of our proposed dataset, significant improvements are observed on the multi-view scanning test dataset proposed in this paper, as indicated in the ``w'' column of Tab.~\ref{tab:5} and Tab.~\ref{tab:4}. In addition, Fig.~\ref{fig:3} and Fig.~\ref{fig:4} also provide visual comparisons of the performance enhancement on CAD models and detailed models after training with the new dataset.

Additionally, we compared the performance of models trained with multi-view point cloud data against the original models on the PCPNet dataset, as shown in Tab.~\ref{tab:2}. The experimental results indicate that incorporating multi-view data for training significantly improves the performance of the PCPNet and DeepFit models. However, for models with stronger learning capabilities, training with additional data that has a significant domain gap results in a slight performance decline on the PCPNet dataset. This suggests an unsettling fact: the high-performing normal estimation models including AdaFit~\cite{zhu2021adafit}, GraphFit~\cite{li2022graphfit} and Hsurf~\cite{li2022hsurf} may actually be in an overfitted state.



\begin{table*}[tb]
\caption{Comparisons of RMSE angle error on the asymmetric Siamese DeepFit with different feature matching weight values.}
\footnotesize
\centering
\begin{tabular}{l ccccccc}
\toprule
\multirow{2}{*}{\textbf{weights}} & \multicolumn{7}{c}{\textbf{$(l_{1},l_{2})$}}\\
\cmidrule(r){2-8} 
&$(0.2,0.0)$  & $(0.4,0.0)$ & $(0.5,0.0)$ & $(0.5,0.1)$ &$(0.5,0.3)$ &$(0.6, 0.0)$ & $(0.7,0.0)$\\
\midrule
None  &6.29  & 6.12 & 6.04  &6.53 &8.71& 6.08  &6.36 \\
\hline
\textbf{Noise} $\sigma$       \\
0.00125 & 8.69 &  9.01 & 9.01  & 9.14&10.50& 8.92&  8.95 \\
0.006   & 16.75  & 16.68 & 16.71& 16.75 & 18.50&16.71 & 16.70\\
0.012   & 22.59  &  22.60 & 22.59 & 22.77 & 26.16&22.59& 22.58\\
\hline
\textbf{Density}   \\
Gradient & 7.13 & 6.76 &6.70 &6.95 & 9.26&6.78&7.18\\
Stripes  & 7.76  & 7.36 &7.31& 7.85 & 10.68&7.44  &7.75\\
\hline
Average  & 11.54 & 11.42 &11.39 & 11.71 &13.97& 11.42 &11.59\\
\bottomrule
\end{tabular}
\label{tab:7}
\end{table*}

\begin{figure*}[htp]
 \centering 
 \includegraphics[width=0.85\textwidth]{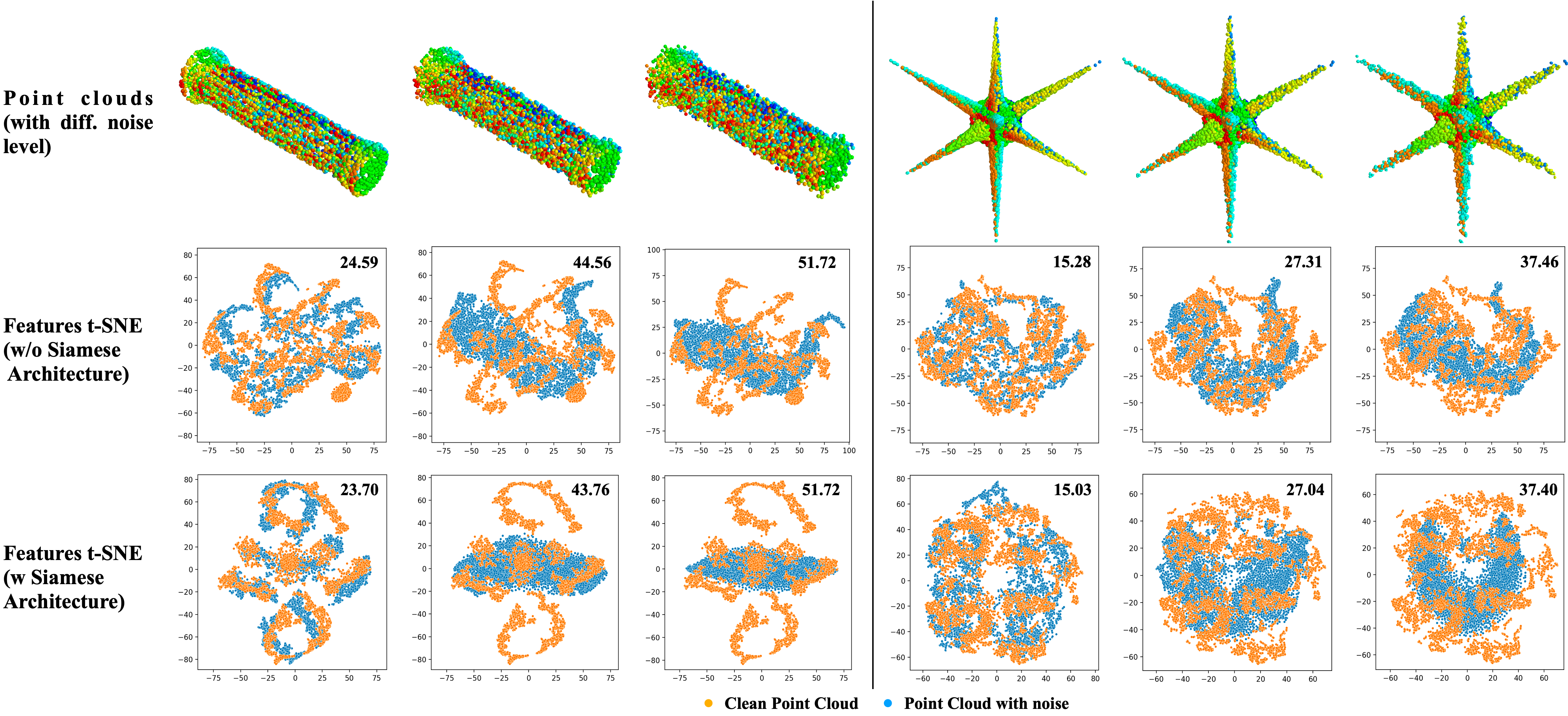}
 \caption{t-SNE visualization of features extracted from the QSTN module.The orange points represent the feature distribution of noise-free point clouds, while the blue points represent the feature distribution of point clouds with different noise levels. The point clouds are from the PCPNet dataset, consisting of models with 5000 sampled points.Each value represents the RMSE angle error of the estimated normal result.}
 \label{fig:8}
\end{figure*}

\subsection{More Analysis for Asymmetrical Siamese Architecture}
\noindent\textbf{Parameter Setting.} Considering that the feature matching weights used in our Asymmetric Siamese Architecture will be utilized to balance the results of normal estimation with the matching degree of noise-free features, which affects the model's generalization ability, the choice of weight values should be carefully discussed and analyzed. In our framework, the global features from the QSTN rotation process and the final latent code are involved in the matching constraints. Therefore, we compared the impact of these two matching losses on the model's accuracy. As shown in Tab.~\ref{tab:7}, our experiments use the DeepFit~\cite{ben2020deepfit} model as the backbone, with the matching learned features in the QSTN module belonging to the $l_{1}$ layer and the final patch's features to the $l_{2}$ layer. The experimental results indicate that the optimal matching weight for the QSTN layer is 0.5. On the other hand, matching the global features of the patch degrades the model's performance. This highlights the critical importance of consistency in the QSTN rotation process.

~

\noindent\textbf{Visualization of Feature Space Drift.} Finally, we visualize the drift in the feature distribution extracted by DeepFit~\cite{ben2020deepfit} and our Asymmetric Siamese DeepFit model using t-SNE. As shown in Fig.~\ref{fig:8}, each example demonstrates the feature drift under low, medium, and high noise levels from left to right. It can be observed that after training with our Asymmetric Siamese Architecture, the feature distribution of the noisy data (blue points) is more aligned with that of the noise-free point clouds, resulting in smaller RMSE angle error. The visualization experiments demonstrate that a more consistent feature distribution contributes to the stability of the model.

\section{Conclusion}
In this study, we introduce an Asymmetric Siamese Network  pipeline for training normal estimation models. Our method enhances the consistency of feature space distribution for multi-scale noisy point clouds compared to the original backbone, leading to improved accuracy for complex models and datasets like FamousShape dataset~\cite{erler2020points2surf}. We also present a larger and more challenging multi-view dataset than PCPNet~\cite{guerrero2018pcpnet} and investigate the domain gap issue between datasets. However, improving the stability of normal estimation models across diverse datasets remains a challenging research problem for future studies.

\bibliographystyle{IEEEtran}
\bibliography{sn-bibliography}

\section*{Biography} 

\vspace{-50pt}
\begin{IEEEbiography}[{\includegraphics[width=1in,height=1.25in,clip,keepaspectratio]{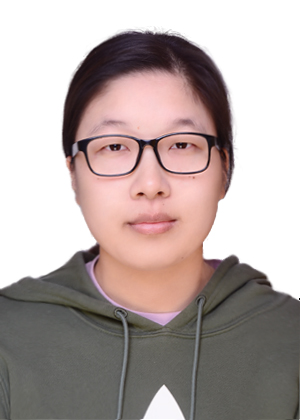}}]{Wei Jin} received her B.S. and M.D. degrees in computational mathematics from Dalian University of Technology, China, in 2012 and 2015, respectively. She is currently a lecturer at Dalian Neusoft University of Information, China. Her research interests include computer graphics, image processing, and machine learning.\end{IEEEbiography}

\vspace{-50pt}
\begin{IEEEbiography}[{\includegraphics[width=1in,height=1.25in,clip,keepaspectratio]{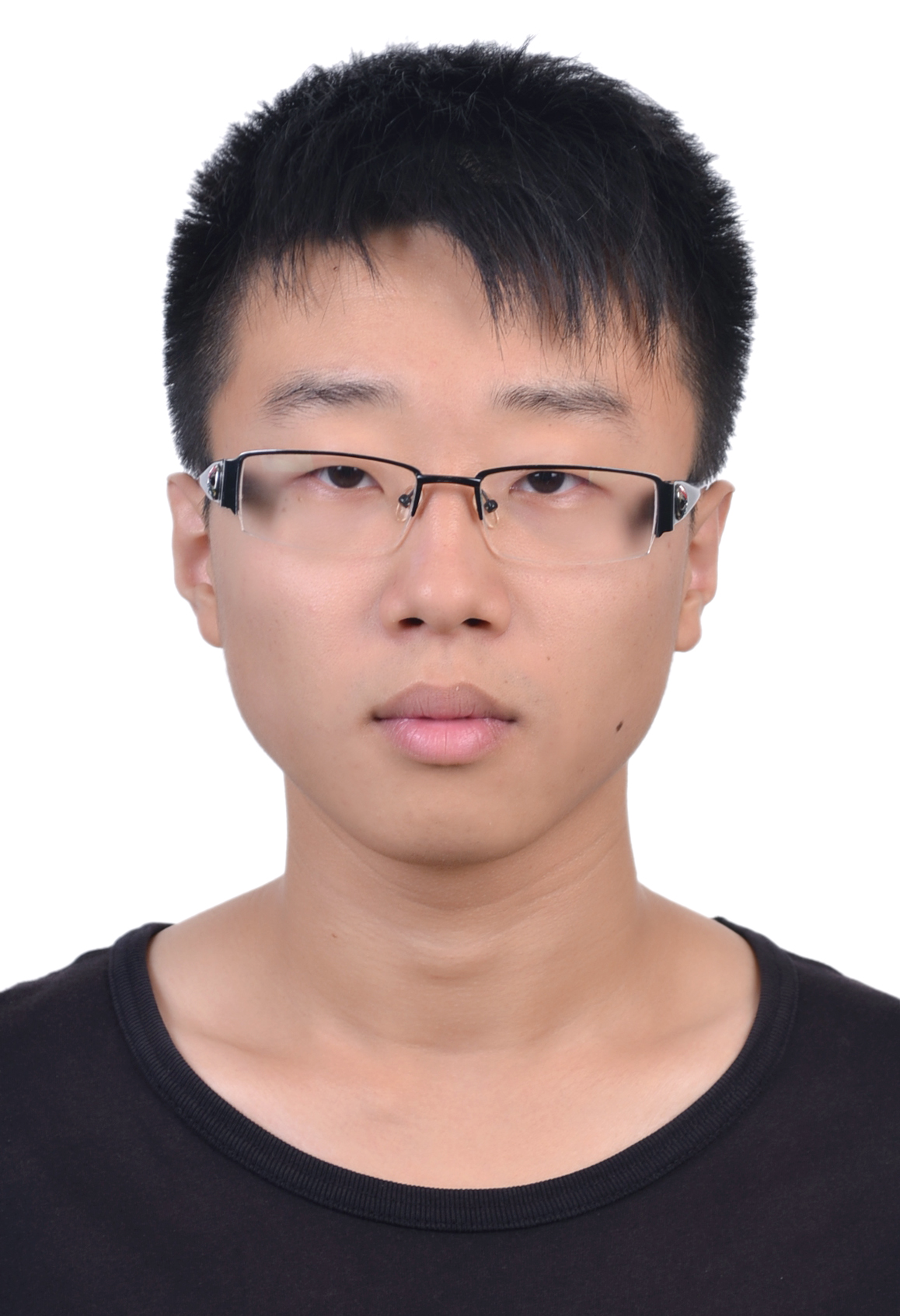}}]{Jun Zhou} received his B.S. and Ph.D. degrees in computational mathematics from Dalian University of Technology, China, in 2013 and 2020, respectively. He is currently a lecturer in the College of Information Science and Technology at Dalian Maritime University, China. His research interests include computer graphics, image processing, and machine learning.\end{IEEEbiography}



\vspace{-50pt}
\begin{IEEEbiography}[{\includegraphics[width=1in,height=1.25in,clip,keepaspectratio]{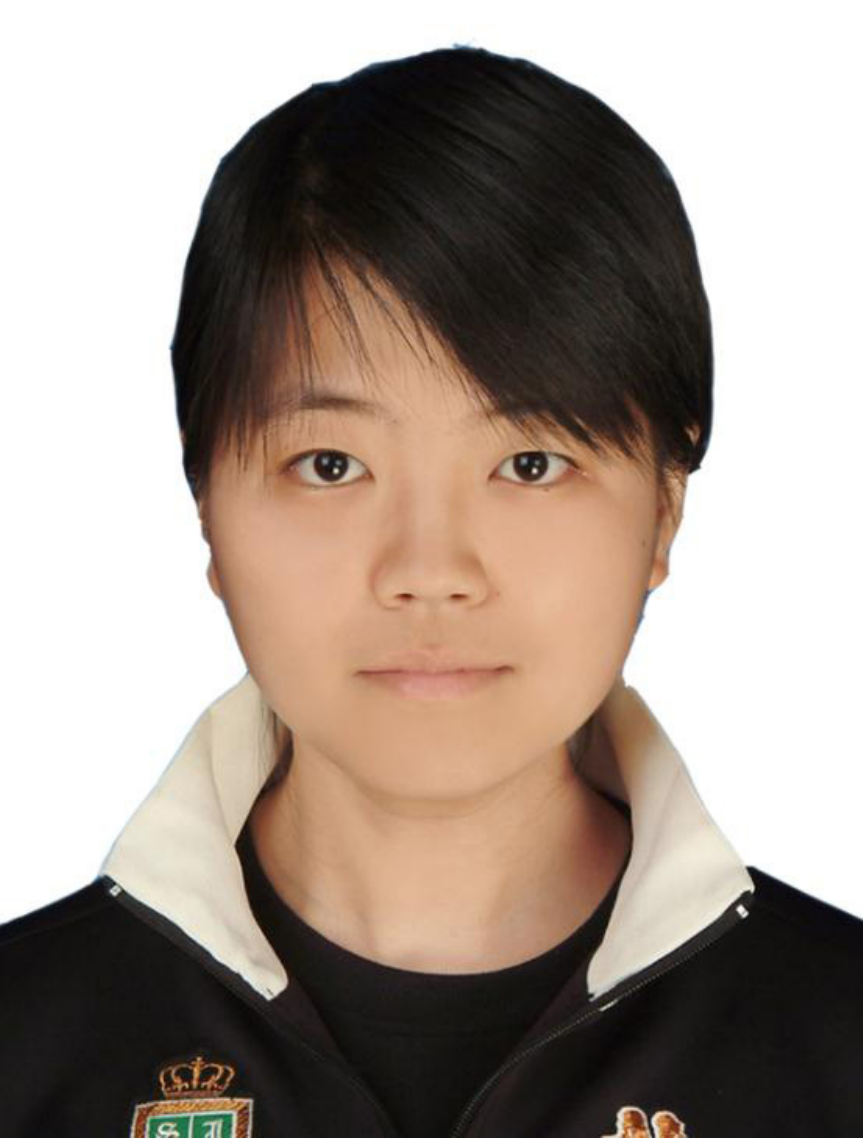}}]{Nannan Li} is an associate professor in School of Information Science and Technology at Dalian  Maritime University. She received her B.S. and M.D-Ph.D  in Computational  Mathematics at Dalian University of Technology. Her  research interests include computer graphics, differential geometry analysis and processing, computer
 aided geometric design and machine learning.\end{IEEEbiography}

\vspace{-50pt}
\begin{IEEEbiography}[{\includegraphics[width=1in,height=1.25in,clip,keepaspectratio]{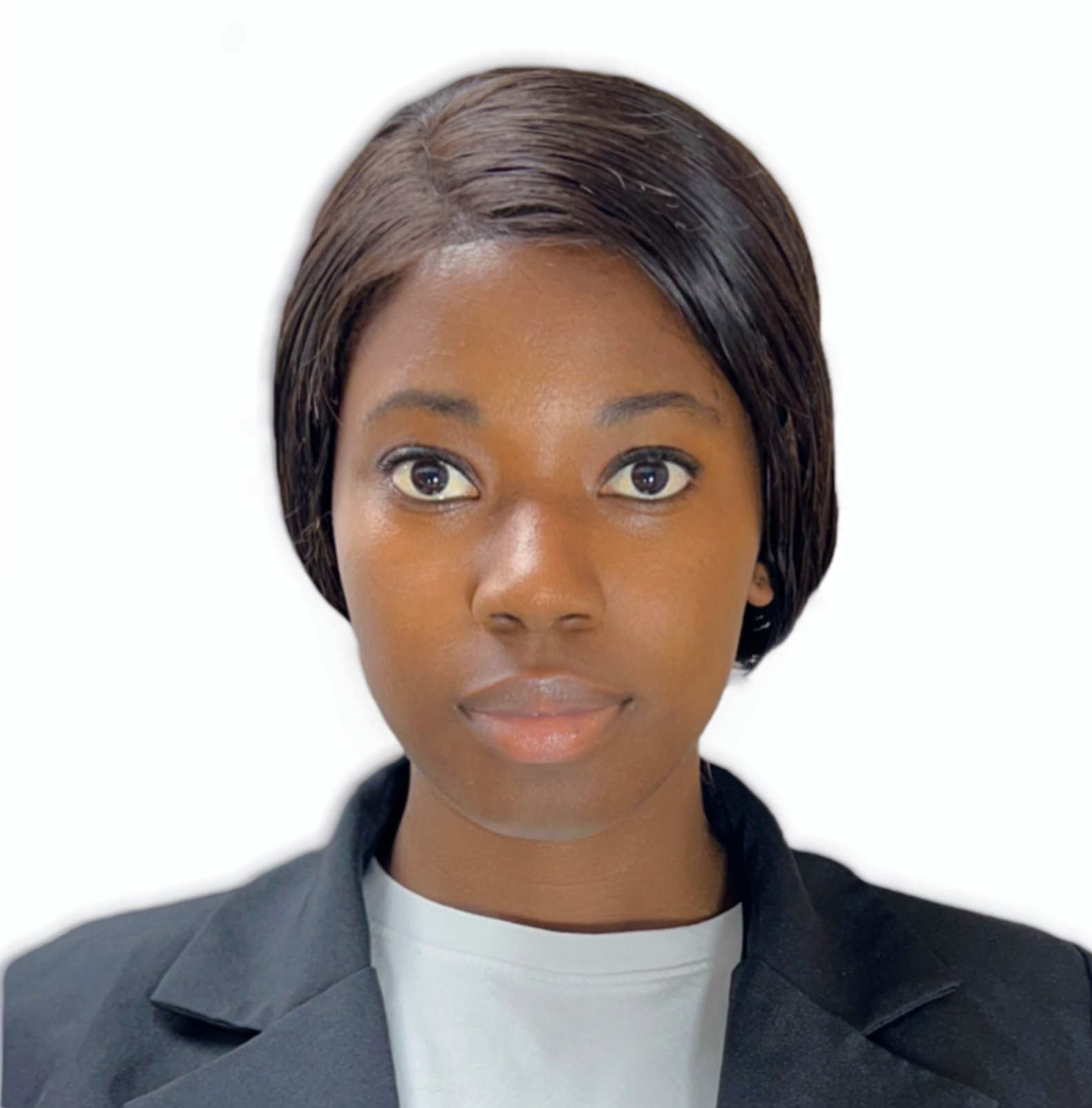}}]{Haba Madeline} holds a Bachelor's Degree in Computer Science and Technology from Shenyang University of Technology, awarded in 2022. She is currently pursuing a Master’s Degree in Software Engineering at Dalian Maritime University. Her research interests include Computer Vision and Pattern Recognition.\end{IEEEbiography}

\vspace{-50pt}
\begin{IEEEbiography}[{\includegraphics[width=1in,height=1.25in,clip,keepaspectratio]{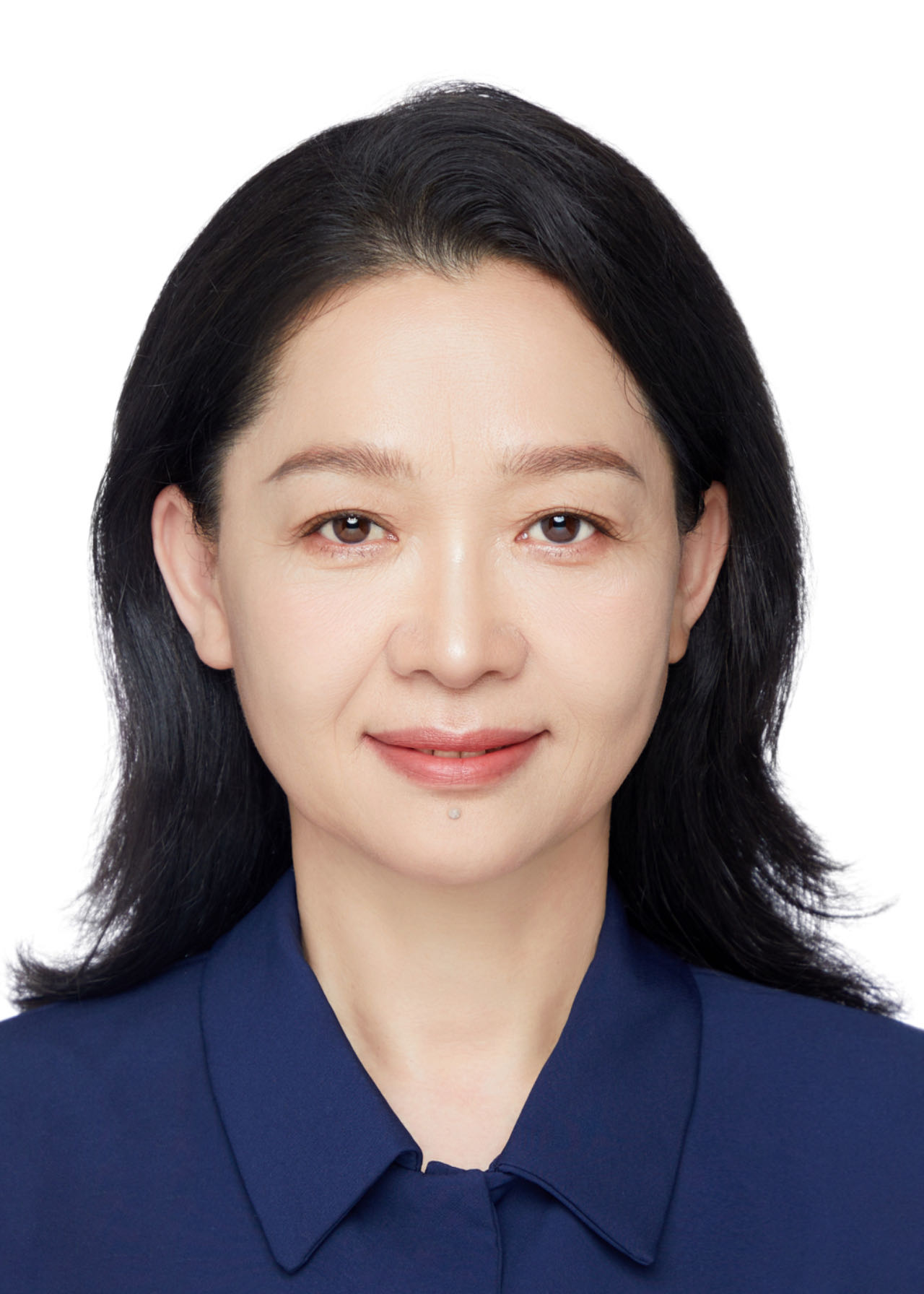}}]{Xiuping Liu} received the BSc degree from Jilin University, China, in 1990, and the PhD degree from the Dalian University of Technology, China, in 1999, respectively. She is a professor with the Dalian University of Technology. Between 1999 and 2001, she conducted research as a postdoctoral scholar in the School of Mathematics, Sun Yat-sen University, China. Her research interests include shape modeling and analyzing.\end{IEEEbiography}

\end{document}